\documentclass[runningheads]{llncs}
\usepackage{graphicx}
\usepackage{amsmath,amssymb} % define this before the line numbering.
\usepackage{color}
\usepackage{hyperref}
\usepackage{subfigure}
\usepackage[table]{xcolor}
\usepackage{multirow}
\usepackage{caption}
\usepackage{url}
\usepackage{doi}
\usepackage[misc]{ifsym}

\begin{document}
%===========================================================

\title{DSNet: Deep and Shallow Feature Learning for Efficient Visual Tracking}  % Replace your paper's title here
\titlerunning{DSNet: Deep and Shallow Feature Learning for Efficient Visual Tracking} % Replace an abstracted version of your paper's title here

%===========================================================

\author{Qiangqiang Wu\and %\inst{1} \orcidID{0000-1111-2222-3333} 
Yan Yan \and %\inst{1} \orcidID{1111-2222-3333-4444}
Yanjie Liang \and %\inst{1} \orcidID{2222--3333-4444-5555}
Yi Liu \and %\inst{1} 
Hanzi Wang\thanks{The corresponding author.}} %\inst{1}  \inst{(}\Letter\inst{)} {$^{\textrm{(\Letter)}}$}
%
%Please include author names in full in the paper, 
%If any authors have names that can be parsed into FirstName LastName in multiple ways, please include the correct parsing, in a comment to the volume editors:
%\index{Lastnames, Firstnames}

\authorrunning{Q. Wu et al.} % A shorter version of authors' name

%===========================================================

\institute{Fujian Key Laboratory of Sensing and Computing for Smart City, \\
School of Information Science and Engineering, Xiamen University, Xiamen, China
\email{\{qiangwu, liuyitan\}@stu.xmu.edu.cn}, \email{yanjieliang@yeah.net}, \\ \email{\{yanyan, hanzi.wang\}@xmu.edu.cn}
}

%\institute{Princeton University, Princeton NJ 08544, USA \and
%Springer Heidelberg, Tiergartenstr. 17, 69121 Heidelberg, Germany

\maketitle

%===========================================================
\begin{abstract}
In recent years, Discriminative Correlation Filter (DCF) based tracking methods have achieved great success in visual tracking. However, the multi-resolution convolutional feature maps trained from other tasks like image classification, cannot be naturally used in the conventional DCF formulation. Furthermore, these high-dimensional feature maps significantly increase the tracking complexity and thus limit the tracking speed. In this paper, we present a deep and shallow feature learning network, namely DSNet, to learn the multi-level same-resolution compressed (MSC) features for efficient online tracking, in an end-to-end offline manner. Specifically, the proposed DSNet compresses multi-level convolutional features to uniform spatial resolution features. The learned MSC features effectively encode both appearance and semantic information of objects in the same-resolution feature maps, thus enabling an elegant combination of the MSC features with any DCF-based methods. Additionally, a channel reliability measurement (CRM) method is presented to further refine the learned MSC features. We demonstrate the effectiveness of the MSC features learned from the proposed DSNet on two DCF tracking frameworks: the basic DCF framework and the continuous convolution operator framework. Extensive experiments show that the learned MSC features have the appealing advantage of allowing the equipped DCF-based tracking methods to perform favorably against the state-of-the-art methods while running at high frame rates. 

\keywords{Visual tracking  \and Correlation filter \and Deep neural network.}
\end{abstract}
%===========================================================

\section{Introduction}
Given the initial state of a target at the first frame, generic visual object tracking is to accurately and efficiently estimate the trajectory of the target at subsequent frames. In recent years, Discriminative Correlation Filter (DCF) based tracking methods have shown excellent performance on canonical object tracking benchmarks \cite{OTB50,OTB100}. The key reasons to their success are the mechanism of enlarging training data by including all shifted samples of a given sample, and the efficiency of DCF by solving the ridge regression problem in the frequency domain. 

Features play an important role in designing a high-performance tracking method. In recent years, significant progress has been made in exploiting discriminative features for DCFs. For example, hand-crafted features like HOG \cite{HOG}, Color Names \cite{CN} or the combinations of these features, are commonly employed by DCFs for online object tracking. Despite the fast tracking speed achieved by these methods, they usually cannot obtain high tracking accuracy due to the less discriminative features they use. Recently, the outstanding success of deep convolutional neural networks (CNNs) has been made in a variety of computer vision tasks \cite{KCF,SSD,Deconv}. Inspired by the success of CNNs, the visual tracking community has exploited the advantages of CNNs, and shown that deep convolutional features trained from other tasks like image classification, are also applicable for the visual tracking task \cite{HDT}. On one hand, the deep features extracted from the shallow convolutional layers, which provide high spatial resolution, are more helpful for the accurate localization of the object. On the other hand, the deep features extracted from the deeper layers encode the semantic information and are more robust to target appearance variations (e.g., deformation, rotation and motion blur). The combination of these two types of features shows excellent tracking performance in both locating the target accurately and modeling the target appearance variations online. However, the conventional DCF formulation is limited to single-resolution feature maps. Deep and shallow features (i.e., multi-resolution feature maps) cannot be naturally used in the conventional DCF framework. Thus, how to effectively fuse the deep and shallow features in the DCF framework is still an open and challenging problem.  

\begin{figure*}[!tp]
\begin{center}
   \includegraphics[width=1.02\linewidth]{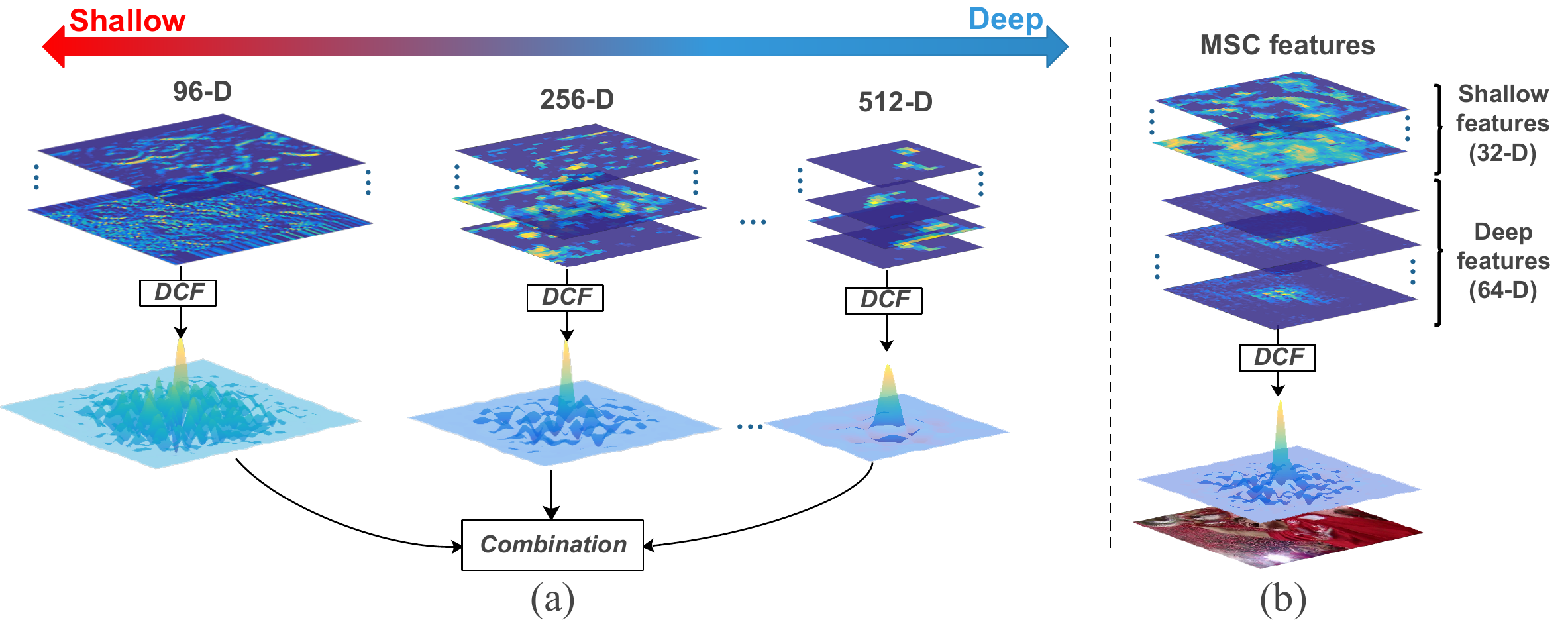} %example.pdf
\end{center}
%\vspace{-0.6cm}
 \caption{Comparison between (a) the DCF-based tracking methods \cite{CCOT,HCF} with deep convolutional features trained from the image classification task and (b) the DCF-based tracking method with our MSC features. 
 %Different from the typical deep convolutional features, our MSC features incorporate both deep and shallow features of objects but with the same spatial resolution, thus enabling to be naturally fused in any DCF-based tracking methods without any online learning formulations.
 }
% the best target window is selected by the proposed two experts and is used to correct the 
% In order to effectively evaluate these candidate results, the proposed on-line expert mechanism (section 3.3) is used to evaluate these candidate results. Specifically, the D-expert and the C-expert are activated to filter and rank all %these k candidate results jointly, and the selected confident result will be used to correct the deep correlation filter. Finally, the correlation filter can track the target well again in the 391-th frame.}
\label{Fig:example}
\end{figure*}

Recently, several works have been developed to fuse multi-resolution feature maps in the DCF framework \cite{ECO,CCOT}. A straightforward strategy is to explicitly resample both deep and shallow features from different spatial resolutions to the same resolution. However, such a strategy introduces artifacts, which severely limit the tracking performance. To overcome the above issue, an online learning formulation is presented in C-COT \cite{CCOT} to integrate multi-resolution feature maps. Despite the promising performance achieved by C-COT, it still has several limitations: (1) The online learning formulation is time-consuming due to the high-dimensional deep features. (2) In order to fuse multi-resolution feature maps, multiple DCFs need to be trained. (3) The method employs the deep features trained from other tasks like image classification. These features are not specifically designed for visual tracking and may limit the tracking performance. 

In this paper, instead of designing an online learning formulation to integrate deep and shallow features (i.e., multi-resolution feature maps) in the DCF framework, we propose to learn multi-layer same-resolution compressed (MSC) features in an end-to-end offline manner for efficient online tracking. To achieve this, a deep and shallow feature learning network architecture (called as DSNet) is developed in this paper. The proposed DSNet aggregates multi-level convolutional features and integrates them into the same-resolution feature maps. In the training stage, a correlation filter layer is added in DSNet, enabling to learn the discriminative MSC features for visual tracking. In the test stage, DSNet acts as a feature extractor without relying on the time-consuming online fine-tuning step. In general, our MSC features learned by the proposed DSNet have the following characteristics: 

(1) MSC features effectively incorporate both the deep and shallow features of objects but with the same spatial resolution. This enables MSC features to be naturally fused in any DCF-based tracking methods without using any online combination strategies.

(2) Due to the low-dimension and uniform spatial resolution of MSC features, the tracking model complexity can be significantly decreased (see Fig. \ref{Fig:example}). Generally, our MSC features can be naturally incorporated by a single DCF instead of multiple DCFs, thus leading to highly efficient online object tracking.
%multi-layer trained DCFs to model multi-resolution feature maps are no needed

%Given a video sequence, 
%In addition, based on the observation that several feature channels have small channel reliability scores (see the definition in Eq. (\ref{TAR})), we further 

%To further refine the learned MSC features, an online channel reliability measurement method is presented. Specifically, based on the observation that several feature channels have small channel reliability scores (see the definition in Eq. (\ref{TAR})), which indicates that these feature channels are more sensitive to the background regions rather than the target regions. Therefore, to achieve better tracking performance, only the feature channels with high channel reliability scores will be reserved to perform online tracking. 

To demonstrate the effectiveness of MSC features learned by the proposed DSNet, we incorporate the MSC features into two state-of-the-art tracking frameworks: the basic DCF framework \cite{KCF} and the continuous convolution operator framework \cite{CCOT}, namely MSC-DCF and MSC-CCO, respectively. Experiments demonstrate that our MSC features have the important advantage of allowing the equipped MSC-DCF and MSC-CCO methods to perform favorably against the state-of-the-art methods at high frame rates.

%We demonstrate the effectiveness of our MSC features on two tracking frameworks: the basic DCF-based framework \cite{KCF} and the continuous convolution operator framework \cite{CCOT}. Experiments show that our MSC features have the important benefit of allowing the equipped CF-based tracking methods to perform favorably against the state-of-the-art methods at high frame rates.

In summary, this paper has the following contributions:

(1) We propose a deep and shallow feature learning network architecture, namely DSNet, enabling to learn multi-level same-resolution compressed (MSC) features for efficient online object tracking in an end-to-end offline manner.

(2) Based on the observation that several feature channels have low channel reliability scores, an online channel reliability measurement (CRM) method is presented to further refine the learned MSC features.

%To further refine the learned MSC features in the online tracking stage, a channel reliability measurement method is presented.

(3) We show that our MSC features are applicable for any CF-based tracking methods. Based on the MSC features, two MSC-trackers (MSC-DCF and MSC-CCO) are presented. Experiments demonstrate that the presented trackers can achieve favorable performance while running at high frame rates.

\section{Related Work}
In this section, we give a brief review to the methods closely related to our work.\\

\noindent \textbf{Correlation filter tracking.} Correlation filter (CF) based tracking methods \cite{DSST,CCOT,KCF} have attracted considerable attention due to their computational efficiency and favorable tracking performance. For example, MOOSE \cite{MOOSE} is the initially proposed CF-based tracking method, which uses grayscale images to train the regression model.  KCF \cite{KCF} further extends MOOSE by using multi-channel features and mapping the input features to a kernel space. Staple \cite{Staple} combines both HOG and Color Name features in a CF framework. Despite the fast tracking speed of their methods, the employed hand-crafted features are still not discriminative enough to handle different challenges. To overcome this problem, several deep feature based tracking methods have been proposed. For example, CF2 \cite{HCF} and DeepSRDCF \cite{DeepSRDCF} employ the convolutional features extracted from VGGNet \cite{VGG-M}. HDT \cite{HDT} merges multiple CFs trained on the different layers of VGGNet. In \cite{CCOT}, an online learning formulation of convolutional features is developed on the spatial domain. ECO \cite{ECO} further alleviates the over-fitting problem in \cite{CCOT}, and decreases the computational complexity. Despite significant improvements made by these methods, they still suffer from the problems of low tracking speed and less discriminative deep features. To alleviate these problems, in this work, we propose to learn the multi-layer same-resolution compressed (MSC) features in an end-to-end offline manner. The learned MSC features are specifically designed for visual tracking, and they can be easily incorporated into any CF-based tracking methods without using the time-consuming online fine-tuning steps. 
 \\

\noindent \textbf{Feature representation learning.} Feature representation is the core of many computer vision tasks, including semantic segmentation \cite{Deconv}, object detection \cite{YOLO} and object tracking \cite{Staple}. For the object detection task, many works on feature learning have been proposed. For example, in R-CNN \cite{R-CNN}, a region proposal based CNN is proposed to learn better feature representations in an end-to-end manner. Due to the learned discriminative features, R-CNN significantly outperforms other detection methods. In addition, SSD \cite{SSD} and HyperNer \cite{HyperNet} combine multi-layer convolution features to further improve the detection performance. For the visual tracking task, CFNet \cite{CFNet} firstly proposes to add a correlation filter layer in a CNN architecture, thus enabling to learn more discriminative features for CF-based methods. The similar feature learning method is also introduced in DCFNet \cite{DCFNet}. Despite the success of these methods, these methods only focus on learning single-layer convolutional features, which may limit their performance. To encode both the low-level and high-level information of objects, we propose to learn multi-layer convolutional features in an end-to-end manner for visual tracking. 
% several detection methods have exploited to leverage multi-layer convolution features. 

\section{Proposed DSNet for Feature Learning}
In this section, we firstly introduce the proposed DSNet framework. Secondly, the feature learning of MSC features is presented. Thirdly, we detail the feature extraction step of MSC features. Finally, we introduce the  channel reliability measurement (CRM) method to further refine the learned MSC features.

\subsection{DSNet Framework} %DSNet framework
\begin{figure*}[!tp]
\begin{center}
   \includegraphics[width=1.02\linewidth]{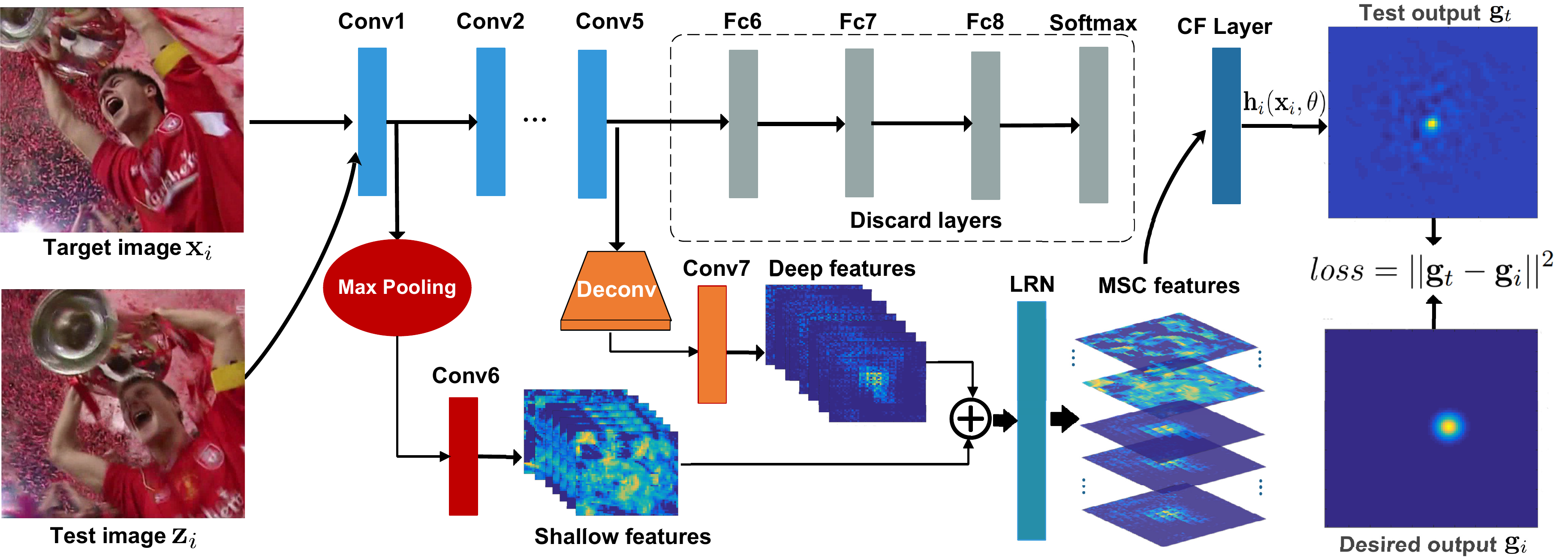} %overall7.eps
\end{center}
%\vspace{-0.6cm}
 \caption{Overall architecture of the proposed deep and shallow feature learning network (DSNet).}
\label{network}
\end{figure*}

The proposed DSNet framework is illustrated in Fig. \ref{network}. As can be seen, DSNet mainly consists of three parts: a backbone network (the blue part), a shallow feature extraction branch (the red part) and a deep feature extraction branch (the orange part). The backbone network is a pre-trained image classification network, which can be any classification CNNs, such as AlexNet \cite{Alexnet} and ResNet \cite{Resnet}. In this work, we select the imagenet-vgg-2048 network \cite{VGG-M} as our backbone network. In the shallow and deep feature extraction branches, in order to combine multi-layer multi-resolution feature maps at the same spatial resolution, a max pooling layer (a 7$\times$7 kernel with a stride of 2) and a deconvolution layer \cite{Deconv} (a 4$\times$4 kernel with a stride of 4) are added to perform downsampling and upsampling, respectively. Then, the Conv6 and Conv7 layers (i.e., the 1$\times$1 convolutional layers) are employed to compress the shallow and deep features, respectively. Moreover, a local response normalization (LRN) layer \cite{Alexnet}  is employed to normalize these features. Finally, we concatenate the normalized features to a single output cube, and obtain our multi-layer same-resolution compressed (MSC) features. In order to effectively train the MSC features, a correlation filter is interpreted as a differentiable layer, which is added at the last layer of DSNet in the training stage.

\subsection{End-to-end Feature Learning}
%Recently, the DCF-based tracking methods \cite{HCF,HDT} with the deep convolutional features have achieved great success. However, the deep features used in these methods are trained from other tasks like image classification. Thus, they are not well-designed for the task of visual object tracking, which may lead to suboptimal tracking results. To alleviate this problem, in the training stage, we add a correlation filter layer (see Fig. \ref{network}) in the proposed DSNet to perform the end-to-end representation learning of our MSC features in an offline manner. 
%To effectively train DSNet, a set of triplet training samples is generated. 
In order to effectively train the proposed DSNet and make the learned MSC features suitable for correlation filter tracking, we add a correlation filter layer (see Fig. \ref{network}) in the proposed DSNet to perform the end-to-end MSC feature representation learning in an offline manner. 

In the training stage, a set of triplet training samples is generated. Let $T=\{\mathbf{x}_{i}, \mathbf{z}_{i}, \mathbf{g}_{i}\}$ be a triplet, where $\mathbf{x}_{i}$ is the target image patch including the centered target, $\mathbf{z}_{i}$ is the test image patch which contains the non-centered target, and $\mathbf{g}_{i}$ is the desired Gaussian distribution centered at the target center position according to $\mathbf{z}_{i}$. Given a  batch size of $N$ triplet training samples, the cost function is formulated as:
\begin{equation}\label{cost_function}
  \mathcal{L}(\mathbf{\theta}) = \sum_{i=1}^{N}{||\sum_{l=1}^{D}{\mathbf{h}_{i}^{l}(\mathbf{\theta})} \ast \varphi^{l}(\mathbf{z}_{i}, \mathbf{\theta})-\mathbf{g}_{i}||^2},
\end{equation}
where
\begin{equation}\label{cost_function2}
\mathbf{h}_{i}^{l}(\mathbf{\theta}) = \mathcal{F}^{-1}(\frac{\hat{\varphi}^{l}(\mathbf{x}_{i}, \mathbf{\theta}) \odot \hat{\mathbf{g}_{i}}^{\ast}}{\sum_{k=1}^{D}{\hat{\varphi}^{k}(\mathbf{x}_{i}, \mathbf{\theta}) \odot (\hat{\varphi}^{k}(\mathbf{x}_{i}, \mathbf{\theta}))^{\ast}+\lambda}})
\end{equation}
%\begin{equation}\label{cost_function}
%\begin{split}
%&  \mathcal{L}(\mathbf{\theta}) = \sum_{i=1}^{N}{||\sum_{l=1}^{D}{\mathbf{h}_{i}^{l}(\mathbf{\theta})} \ast \varphi^{l}(\mathbf{z}_{i}, \mathbf{\theta})-\mathbf{g}_{i}||^2}  \\ 
%&  \text{where~} \quad \mathbf{h}_{i}^{l}(\mathbf{\theta}) = \mathcal{F}^{-1}(\frac{\hat{\varphi}^{l}(\mathbf{x}_{i}, \mathbf{\theta}) \odot \hat{\mathbf{g}_{i}}^{\ast}}{\sum_{k=1}^{D}{\hat{\varphi}^{k}(\mathbf{x}_{i}, \mathbf{\theta}) \odot (\hat{\varphi}^{k}(\mathbf{x}_{i}, \mathbf{\theta}))^{\ast}+\lambda}})
%\end{split}
%\end{equation}
and $\mathbf{h}_{i}^{l}(\mathbf{\theta})$ is the desired correlation filter for the $l$-th channel feature map,  $\mathbf{\theta}$ refers to the parameters of our DSNet, $\varphi^{l}(\mathbf{x}_{i}, \mathbf{\theta})$ is the extracted features of $l$-th channel with the parameters $\mathbf{\theta}$ corresponding to the input $\mathbf{x}_{i}$, and $\lambda$ is a regularization parameter. Furthermore, $\mathcal{F}^{-1}$ denotes the inverse Fourier transform, $\ast$ is the circular correlation operation, $D$ represents the channel numbers, $\odot$, $\hat{}$ and ${}^{\ast}$ denote the Hadamard product, discrete Fourier transform and complex conjugation, respectively. By applying the multivariable chain rule, the derivation of the loss function in Eq. (\ref{cost_function}) can be rewritten as:
%According to the multivariable chain rule, the derivation of the loss function in (\ref{cost_function}) can be rewritten as:
\begin{equation}\label{chain_loss}
   {\frac{\partial \mathcal{L}}{\partial \mathbf{\theta}} = \sum_{l}\frac{\partial \mathcal{L}}{\partial \varphi^{l}(\mathbf{x}_{i}, \mathbf{\theta})}\frac{\partial \varphi^{l}(\mathbf{x}_{i}, \mathbf{\theta})}{\partial \mathbf{\theta}} + \sum_{l}\frac{\partial \mathcal{L}}{\partial \varphi^{l}(\mathbf{z}_{i}, \mathbf{\theta})}\frac{\partial \varphi^{l}(\mathbf{z}_{i}, \mathbf{\theta})}{\partial \mathbf{\theta}}}.
\end{equation}
Specifically, $\frac{\partial \varphi^{l}(\mathbf{x}_{i}, \mathbf{\theta})}{\partial \mathbf{\theta}}$ and $\frac{\partial \varphi^{l}(\mathbf{z}_{i}, \mathbf{\theta})}{\partial \mathbf{\theta}}$ in the above can be efficiently calculated by recent deep learning toolkits.
According to \cite{CFNet,DCFNet}, the prior two terms ($\frac{\partial \mathcal{L}}{\partial \varphi^{l}(\mathbf{x}_{i}, \mathbf{\theta})}$ and $\frac{\partial \mathcal{L}}{\partial \varphi^{l}(\mathbf{z}_{i}, \mathbf{\theta})}$) in (\ref{chain_loss}) can be formulated as:
\begin{equation}\label{loss_solve_}
  \frac{\partial \mathcal{L}}{\partial \varphi^{l}(\mathbf{x}_{i}, \mathbf{\theta})} = \mathcal{F}^{-1}(\frac{\partial \mathcal{L}}{\partial (\hat{\varphi}^{l}(\mathbf{x}_{i}, \mathbf{\theta}))^{\ast}}+ (\frac{\partial \mathcal{L}}{\partial \hat{\varphi}^{l}(\mathbf{x}_{i}, \mathbf{\theta})})^{\ast}),
\end{equation}
\begin{equation}\label{loss_solve}
  \frac{\partial \mathcal{L}}{\partial \varphi^{l}(\mathbf{z}_{i}, \mathbf{\theta})} = \mathcal{F}^{-1}(\frac{\partial \mathcal{L}}{\partial (\hat{\varphi}^{l}(\mathbf{z}_{i}, \mathbf{\theta}))^{\ast}}).
\end{equation}

%During the offline training process, the gradients in Eq. ($\ref{chain_loss}$) are back-propagated, and the whole network parameters $\mathbf{\theta}$ of DSNet are updated to minimize the loss function in Eq. ($\ref{cost_function}$). 

%In the test stage, our DSNet acts as a feature extractor, which is used to extract robust MSC features for efficient online object tracking.

\subsection{Feature Extraction}
\begin{figure*}[!tp]
\begin{center}
   \includegraphics[width=1.02\linewidth]{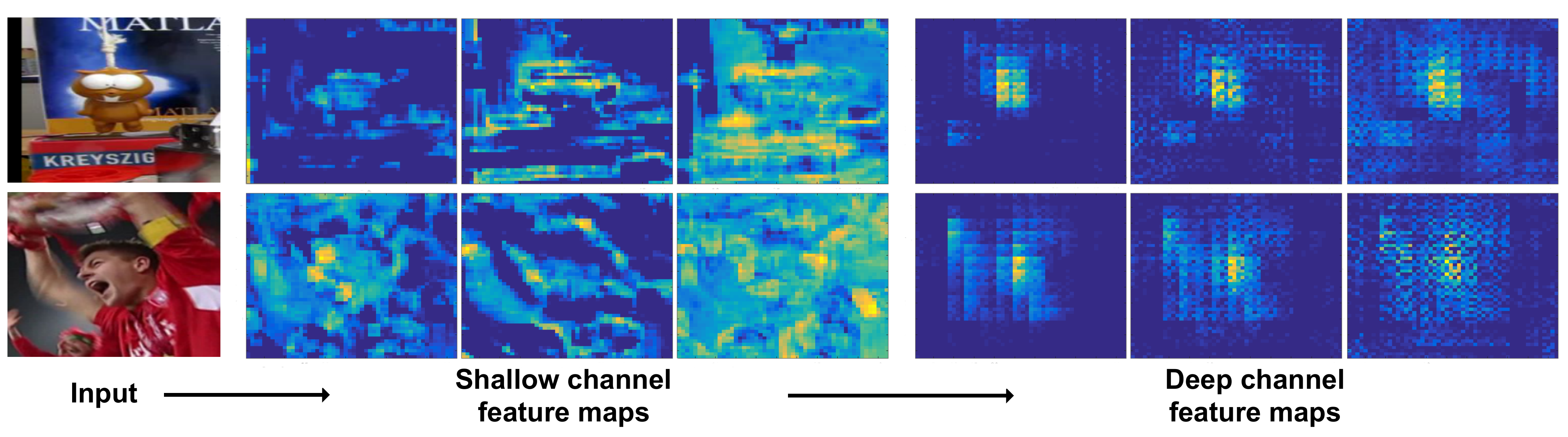} %overall7.eps
\end{center}
%\vspace{-0.6cm}
 \caption{Visualization of the shallow and deep channel feature maps in the learned MSC features. }
\label{features}
\end{figure*}

In the online tracking stage, DSNet acts as a feature extractor, which first extracts multi-layer convolutional feature maps as shown in Fig. \ref{network}. Next, the shallow and deep feature extraction branches in our DSNet aggregate multi-level convolutional feature maps and compress them to the uniform spatial resolution features. Finally, the MSC features are obtained by normalizing the compressed convolutional features.  Specifically, the obtained MSC features (with the size of $52\times52\times96$) consist of two parts of features: shallow convolutional features with the size of $52\times52\times32$ and deep convolutional features with the size of $52\times52\times64$. To better understand the learned MSC features, several channel feature maps of MSC features are visualized in Fig. \ref{features}. As can be seen, the shallow channel feature maps usually capture the detailed information of objects, while the deep channel feature maps generally encode the semantic information. These two types of features can complement each other, and the combination of them is beneficial for online tracking.

\subsection{Channel Reliability Measurement}
%Given a test video, 
%Motivated by the observation that several channel feature maps of MSC features may have small target activations, which indicates that these feature channels are more sensitive to the background regions rather than the target regions.

Several channel feature maps of MSC features may have small target activations, which indicates that these feature channels are more sensitive to the background regions rather than the target regions. To measure the reliability of these channels, the channel-wise ratio of the $l$-th channel is formulated as:

\begin{equation}\label{TAR}
 R^{l} = \frac{||S_{t}^{l}||_{1}}{||S_{e}^{l}||_{1} + \zeta}.
  %TAR_{l}=\frac{\sum_{w_{t}=1}^{W_{t}} \sum_{h_{t}=1}^{H_{t}} \phi^{l}(w_{t}, h_{t})}{\sum_{w_{o}=1}^{W_{o}} \sum_{h_{o}=1}^{H_{o}} \phi^{l}(w_{o}, h_{o})}
\end{equation}
Here, $S_{e}^{l}$ refers to the entire $l$-th channel feature map, $S_{t}^{l}$ is the target region of $S_{e}^{l}$, $\zeta$ is a penalty parameter and ${||\cdot||}_{1}$ is the $l_{1}$ norm. 

The channel-wise ratio shows the ratio of the target responses to the overall responses, however, it cannot fully reflect the channel reliability in some cases. For example, when the background responses are equal to zero, even $||S_{t}^{l}||_{1}$ is a small value, a quite large channel-wise ratio $R^{l}$ will be obtained. To overcome the above problem, $A^{l}$ is defined to measure the target activations of the $l$-th channel feature map:
\begin{equation}\label{E}
\begin{split}
&  A^{l} = 
\begin{cases}
1 &  Z(S_{t}^{l})>W_{t}H_{t}/\eta  \\
0 & \text{otherwise},
\end{cases}
\end{split}
\end{equation}
where
\begin{equation}\label{E2}
Z(S_{t}^{l}) = \sum_{w=1}^{W_{t}} \sum_{h=1}^{H_t} sign(|S_{t}^{l}(w, h)|),
\end{equation}
and $sign()$ is the sign function, $S_{t}^{l}(w, h)$ returns the activation value at the position $(w, h)$ of $S_{t}^{l}$, $W_t$ and $H_t$ are the width and height of $S_{t}^{l}$, respectively. $\eta$ is a penalty parameter that controls the measurement of the target region responses. Finally, the reliability score of the $l$-th feature channel is calculated by:
\begin{equation}\label{TAR}
 C^l = R^{l} \times A^{l}.
  %TAR_{l}=\frac{\sum_{w_{t}=1}^{W_{t}} \sum_{h_{t}=1}^{H_{t}} \phi^{l}(w_{t}, h_{t})}{\sum_{w_{o}=1}^{W_{o}} \sum_{h_{o}=1}^{H_{o}} \phi^{l}(w_{o}, h_{o})}
\end{equation}
%The correlation filter layer is used in several recent proposed methods \cite{}, and it interprets the correlation filters as a differentiable layer in the network, enabling to learn more expressive features for correlation filter tracking.

%Generally, channels with high $C^l$ reflect that they contain more activations from the target regions than the background regions. After obtaining the reliability scores of both deep and shallow feature channels, we sort these channels independently in the descending order. To achieve better tracking performance, the top ranked K$_d$ deep channels and K$_s$ shallow channels will be selected to be used in online tracking. 

Generally, channels with high $C^l$ reflect that they contain more activations from the target regions than the background regions. After obtaining the reliability scores of feature channels, we sort these channels in the descending order. The top ranked $K$ feature channels are selected to perform online tracking. 
%Motivated by the observation that the shallow features are more compact and the deep features may contain several noisy channels, to achieve better tracking performance, we reserve all the shallow channels and select t
%Thus, these channels capture more information of tracked objects and are helpful for online tracking. independently
%all the feature channels, we independently  sort all the feature channels according to $C^l$ in the descending order. To achieve the better tracking performance, the last sorted $K_c$ feature channels will not be used in online object tracking. 

%1. Preliminaries.
%2. Network architecture.
%3. MSC Feature learning.
%4. Channel reliability measurement method.

\section{MSC-Trackers}
In this section, we show how the learned MSC features can be incorporated into different DCF-based tracking frameworks. We select two state-of-the-art tracking frameworks, i.e., the basic DCF framework \cite{KCF} and the continuous convolutional operator framework \cite{CCOT}.   
\subsection{MSC Features for the Basic DCF framework} 
A typical DCF learns a correlation filter $\mathbf{h}^{l}$ by solving a Ridge Regression problem:
\begin{equation}\label{DCF_loss}
 \min_{\mathbf{h}^{l}}  ||\sum_{l=1}^{D} {\mathbf{h}^{l}} \ast \varphi^{l}(\mathbf{x}, \mathbf{\theta})-\mathbf{g}||^{2} + \lambda_{D}\sum_{l=1}^{D}||\mathbf{h}^{l}||^{2},
\end{equation}
where $\varphi^{l}(\mathbf{x}, \mathbf{\theta})$ is the extracted MSC features of $l$-th channel with the DSNet parameters $\mathbf{\theta}$ corresponding to the training image patch $\mathbf{x}$, $\mathbf{g}$ is the desired Gaussian distribution, $\lambda_{D}$ is a regularization parameter that alleviates the overfitting problem. The learned correlation filter $\mathbf{h}^{l}$ can be obtained as \cite{KCF}:
\begin{equation}\label{DCF_solution}
\hat{\mathbf{h}^{l}} = \frac{\hat{\varphi}^{l}(\mathbf{x}, \mathbf{\theta}) \odot \hat{\mathbf{g}}^{\ast}}{\sum_{k=1}^{D}{\hat{\varphi}^{k}(\mathbf{x}, \mathbf{\theta}) \odot (\hat{\varphi}^{k}(\mathbf{x}, \mathbf{\theta}))^{\ast}+\lambda_D}}.
\end{equation}
Given the test image patch $\mathbf{z}$ and the extracted features $\varphi(\mathbf{z}, \mathbf{\theta})$, the online detection process is formulated as:
\begin{equation}\label{Detection}
\mathbf{f} = \mathcal{F}^{-1}(\sum_{l=1}^{D}\hat{\mathbf{h}^{l}}^{\ast} \odot \hat{\varphi}^{l}(\mathbf{z, \mathbf{\theta}})),
\end{equation}
where $\mathbf{f}$ is the response map. The target center position can be estimated by searching the maximum value in $\mathbf{f}$. During the tracking process, at the $(t+1)$-th frame, the numerator $A_{(t+1)}^{l}$ and denominator $B_{(t+1)}^{l}$ in Eq. (\ref{DCF_solution}) are respectively updated by using a moving average strategy with a learning rate $\mu$. Then the correlation filter model at the $(t+1)$-th frame is updated by $\hat{\mathbf{h}^{l}_{(t+1)}} = A_{(t+1)}^{l} / (B_{(t+1)}^{l}+\lambda_{D})$. We use the scale estimation similar to \cite{DCFNet}.
%\begin{equation}\label{updated1}
%A_{(t+1)}^{l} = (1-\mu)A_{t}^{l} + \mu \hat{\varphi}^{l}(\mathbf{x}_{t+1}, \mathbf{\theta}) \odot \hat{\mathbf{g}}^{\ast}
%\end{equation}
%\begin{equation}\label{updated2}
%B_{(t+1)}^{l} = (1-\mu)B_{t}^{l} + \mu {\sum_{k=1}^{d}{\hat{\varphi}^{k}(\mathbf{x}_{t+1}, \mathbf{\theta}) \odot (\hat{\varphi}^{k}(\mathbf{x}_{t+1}, \mathbf{\theta}))^{\ast}}}
%\end{equation}

Note that the conventional DCF framework is restricted to single-resolution feature maps. In comparison, the proposed MSC features can be naturally fused into the DCF framework (see Fig. \ref{Fig:example}). For briefly, this MSC features based DCF tracker is named as MSC-DCF.  % like the other commonly used features, e.g., the HOG descriptor \cite{}. In comparison,Multi-layer deep convolutional features trained from other tasks like the image classification task, cannot be used in this framework without any modifications. 

\subsection{MSC Features for the Continuous Convolution Operator Framework}
The continuous convolution operator is proposed in C-COT \cite{CCOT}. Let $y_{j}$ denote a training sample, which contains $D$ feature channels $y_{j}^{1},y_{j}^{2}, ...,y_{j}^{D}$. $N_D$ is the number of spatial samples in $y_{j}^{d}$. Here, the feature channel $y_{j}^{d}$ can be viewed as a function $y_{j}^{d}[n]$, where $n$ is the discrete spatial variable $n\in\{0,...,N_d-1\}$. Assume that the spatial support of the feature map is the continuous interval $[0,P)\subset \mathbb{R}$. The interpolation operator $J_d$ is formulated as:
\begin{equation}\label{Jd}
J_{d}\{y^{d}\}(p) = \sum_{n=1}^{N_d}y^{d}[n]b_{d}(p-\frac{P}{N_d}n),
\end{equation}
where $b_d$ is the interpolation function, $p$ denotes the location in the image, $p\in[0, P)$. In the continuous formulation, the convolution operator is estimated by a set of convolution filters $f=(f^1, f^2,...,f^D)\in L^{2}(P)$. The convolution operator is defined as:
\begin{equation}\label{conv}
Q_{f}\{y\} = \sum_{d=1}^{D}f^{d}\circledast J_{d}\{y^{d}\}.
\end{equation}
Here, $f^d$ is the continuous filter for the $d$-th channel (see \cite{CCOT} for more details), $\circledast$ is the circular convolution operation: $L^2(T)\times L^2(T)  \rightarrow L^2(T)$. As can be seen, for each interpolated sample $J_{d}\{y^{d}\}(p)$, it is convolved with the corresponding filter $f^d$. At last, the final confidence map is obtained by summing up the convolution responses from all the filters. 
%To train the continuous filter $f$, a set of training samples is needed. Given one training sample $y_j$, let $v_j$ be the desired output of $Q_{f}\{x_j\}$.  The continuous filter $f$ can be calculated by minimizing a cost function, which is defined as (see \cite{} for more details):
%\begin{equation}\label{cost_ccot}
%\min_{f} \sum_{j=1}^{m}\alpha_{j}||Q_{f}{(y_{j})}-v_{j}||^{2} + \omega\sum_{d=1}^{D}||f^{d}||^{2}
%\end{equation}
%where $m$ is the number of training samples, $\alpha_{j}$ is the weight of each training sample, and $\omega$ is a penalty parameter. The minimization problem in the above can be effectively solved by using the Conjugate Gradient method.

In Eqs. (\ref{Jd}) and (\ref{conv}), the interpolation operator $J_d$ and convolution filter $f^d$ are learned for each feature channel. Thus, the high-dimensional convolutional features (e.g., the 608-dimensional features used in CCOT), significantly limit the online tracking speed. In comparison, the learned MSC features have much less channels (i.e., 96), and they can be regarded as one layer convolutional features to be fused in the continuous convolution operator framework without any modifications. We call this MSC features based tracker as MSC-CCO. 

%increase the tracking model complexity and thus 
%including the 96-dimensional \emph{Conv1} and 512-dimensional \emph{Conv5} features extracted from the imagenet-vgg-2048 network 
%Further, as shown in Fig. ??,  MSC features decrease the learned correlation filter models in the 

%This document serves as an example submission. It illustrates the format
%we expect authors to follow when submitting a paper to ACCV. 
%At the same time, it gives details on various aspects of paper submission,
%including preservation of anonymity and how to deal with dual submissions,
%so we advise authors to read this document carefully.
%Do not use any additional Latex macros.

%------------------------------------------------------------------------- 

\section{Experiments}

%two trackers are presented, i.e., a conventional DCF tracker with MSC features (MSC-DCF)  and a continuous convolution operator tracker with MSC features (MSC-CCO). 
\textbf{Implementation Details:} To avoid overfitting, we select the large scale video detection dataset (ILSVRC-2015) \cite{ILSVRC} to train the proposed DSNet. This dataset contains 4417 videos of 30 different objects. We use 3862 videos in this dataset for training and the remaining videos for validation. The triplet training samples $T$ are generated as described in \cite{CFCF}. The proposed DSNet is trained for 200 epochs with a batch size of 16 and an initial learning rate of $1\times 10^{-5}$ by using the SGD solver. The momentum and weight decay are respectively set to $9\times 10^{-1}$ and $5\times10^{-4}$. In the tracking stage, for MSC-DCF, the learning rate $\mu$ and the padding area are respectively set to 0.012 and 1.65. The searching scale number is set to 3. For MSC-CCO, we set the learning rate and the padding area to $ 9.4\times 10^{-3}$ and 3.62, respectively. Similar to \cite{ECO}, except for MSC features, MSC-CCO also employs HOG features, and the MSC features are further compressed to 38-D by using PCA. The other parameters in MSC-DCF and MSC-CCO are respectively set to be the same as in \cite{DCFNet} and \cite{ECO}. For the CRM method, we apply it to refine the deep feature channels in MSC features, where $K$ in MSC-DCF and MSC-CCO are respectively set to 50 and 58. The penalty parameters $\eta$ and $\zeta$ are set to 3 and $1\times10^{-5}$, respectively. In addition, we implement our method on a computer equipped with an Intel 6700K 4.0 GHz CPU and an NVIDIA GTX 1080 GPU. \\

\noindent \textbf{Evaluation Methodology:} Both the distance precision (DP) and overlap success (OS) plots are adopted to evaluate trackers on OTB-2013 \cite{OTB50}, OTB-2015 \cite{OTB100} and OTB-50. We report both the DP rates at the conventional threshold of 20 pixels (DPR) and the OS rates at the threshold of 0.5 (OSR). The Area Under the Curve (AUC) is also used to evaluate the trackers. \\
% The DP plot shows the percentage of the frames, in which the distance error is within a distance threshold. The OS plot depicts the percentage of the frames whose intersection over union (IoU) score is larger than a threshold. \\
%We report both the DP rates at the conventional threshold of 20 pixels (DPR) and the OS rates at the threshold of 0.5 (OSR). The Area Under the Curve (AUC) is also used to evaluate the trackers. \\
%Note that OTB-50 includes the challenging videos, and it is a subset of OTB-2015. 

\noindent \textbf{Comparison Scenarios:} We evaluate the proposed MSC-trackers (MSC-DCF and MSC-CCO) in four experiments. The first experiment is conducted to demonstrate the effectiveness of the learned MSC features by comparing our MSC features with both hand-crafted features and deep features.  At the second experiment, we compare the proposed highly efficient MSC-trackers with the state-of-the-art real-time trackers, which shows the superiority of our MSC-trackers. The third experiment compares our MSC-trackers with the top-performing CF-based trackers with deep features. The last experiment makes an ablation study on MSC-trackers to show the effectiveness of the proposed CRM method.
%the commonly used features including both 

%\vspace{2cm}

%------------------------------------------------------------------------- 
\subsection{Feature Comparison}

\begin{table*}[!tp]
\newcommand{\tabincell}[2]{\begin{tabular}{@{}#1@{}}#2\end{tabular}}
\small
\begin{center}
%\vspace{-0.391cm}
\caption {\label{baseline}Comparison of our MSC features with both the hand-crafted features and deep convolutional features within the DCF framework on OTB-2013. Note that $^{*}$ indicates the GPU speed, otherwise the CPU speed. The {\color{red} first}, {\color{green} second} and  {\color{blue} third} best features are shown in color.}
%\label{table:headings}
\begin{tabular}{c|c|c|c|c|c|c|c}
\hline\noalign{\smallskip}
&  \textbf{\tabincell{ccc}{MSC}} & \tabincell{ccc}{Conv1} &  \tabincell{ccc}{Conv5} & \tabincell{ccc}{HOG}  & \tabincell{ccc}{RGB} & \tabincell{ccc}{HOG+RGB} & \tabincell{ccc}{Conv1+Conv5}\\
\noalign{\smallskip}
\hline
\noalign{\smallskip}

\hline
\multirow{1}{*}{DPR (\%)}
 & {\color{red}83.7} & 76.2 & 69.4 & 74.4 & 47.4 & {\color{blue}77.7} & {\color{green}78.5}\\
 \hline
 \multirow{1}{*}{OSR (\%)}
 &{\color{red}78.3} & 72.6 & 53.7 & 73.5 & 38.7 & {\color{blue}75.2}  & {\color{green}75.7}  \\
 \hline
  \multirow{1}{*}{Feat. Size.}
 &$52\times52$ &  $109\times109$ & $13\times13$ & $52\times52$ & $52\times52$ & $52\times52$ & $109\times109$   \\
 \hline
 \multirow{1}{*}{Feat. Dimen.}
 &96 &  96 & 512 & 32 & 3 & 35 & 608   \\
  \hline
\multirow{1}{*}{Avg. FPS}
 &${\color{green}68.5}^*$ & $51.0^*$ & ${\color{blue}58.1}^*$ & 56.0 & {\color{red}328.1} & 49.6 & $2.6^*$   \\
\hline
%\multirow{1}{*}{Implemantation}
%& & M C & M C & M C & M C & M & M & M & M C & M C & M & M C & M C\\

\hline
\end{tabular}
\end{center}
\end{table*}

To demonstrate the effectiveness of the learned MSC features, we compare the MSC features with different commonly used features. For fair comparison, all the compared features are incorporated into the same tracking framework, i.e., the DCF framework described in Section 4.1 for this experiment. We compare MSC features with raw RGB pixels (RGB), HOG \cite{HOG}, the first (Conv1) and fifth (Conv5) layer convolutional features in imagenet-vgg-2048, the combination of HOG and RGB (HOG+RGB) features and the combination of Conv1 and Conv5 (Conv1+Conv5) features. In order to combine the Conv1 and Conv5 features, the bilinear interpolation method is employed to upsample the Conv5 features to the same size as the Conv1 features.
%the fifth convolutional layer (Conv5) features in imagenet-vgg-2048

Table \ref{baseline} shows a comparison of our MSC features with different types of features on OTB-2013. As can be seen, our MSC features achieve the best DPR (83.7\%) and OSR (78.3\%) on OTB-2013, significantly outperforming both the hand-crafted features and deep convolutional features with large margins. In particular, the Conv1 and Conv5 features are extracted from the imagenet-vgg-2048 network, which is also the backbone network in DSNet. Despite the similar network architecture, our MSC features improve 7.5\% and 14.4\% of the DPRs obtained by the Conv1 and Conv5 features on OTB-2013, respectively. Compared with the Conv1 and Conv5 features, the combination Conv1+Conv5 features have more feature channels (608) and achieve better performance, with a DPR of 78.5\%. Although much more feature channels are included in the Conv1+Conv5 features, our MSC features still provide improved performance, with a DPR of 83.7\%, while achieving the fast tracking speed of 68.5 FPS, which is about 26 times faster than the Conv1+Conv5 features. This empirically shows that the MSC features are compact and can lead to highly efficient online tracking.

\subsection{Comparison with Real-Time Trackers}
\begin{figure*}[!tp]
    %\vspace{-0.1cm}
 \begin{center}
 \subfigure{
\begin{minipage}[b]{0.487\linewidth}
  \centerline{\includegraphics[width=5.98cm, height=4.5cm]{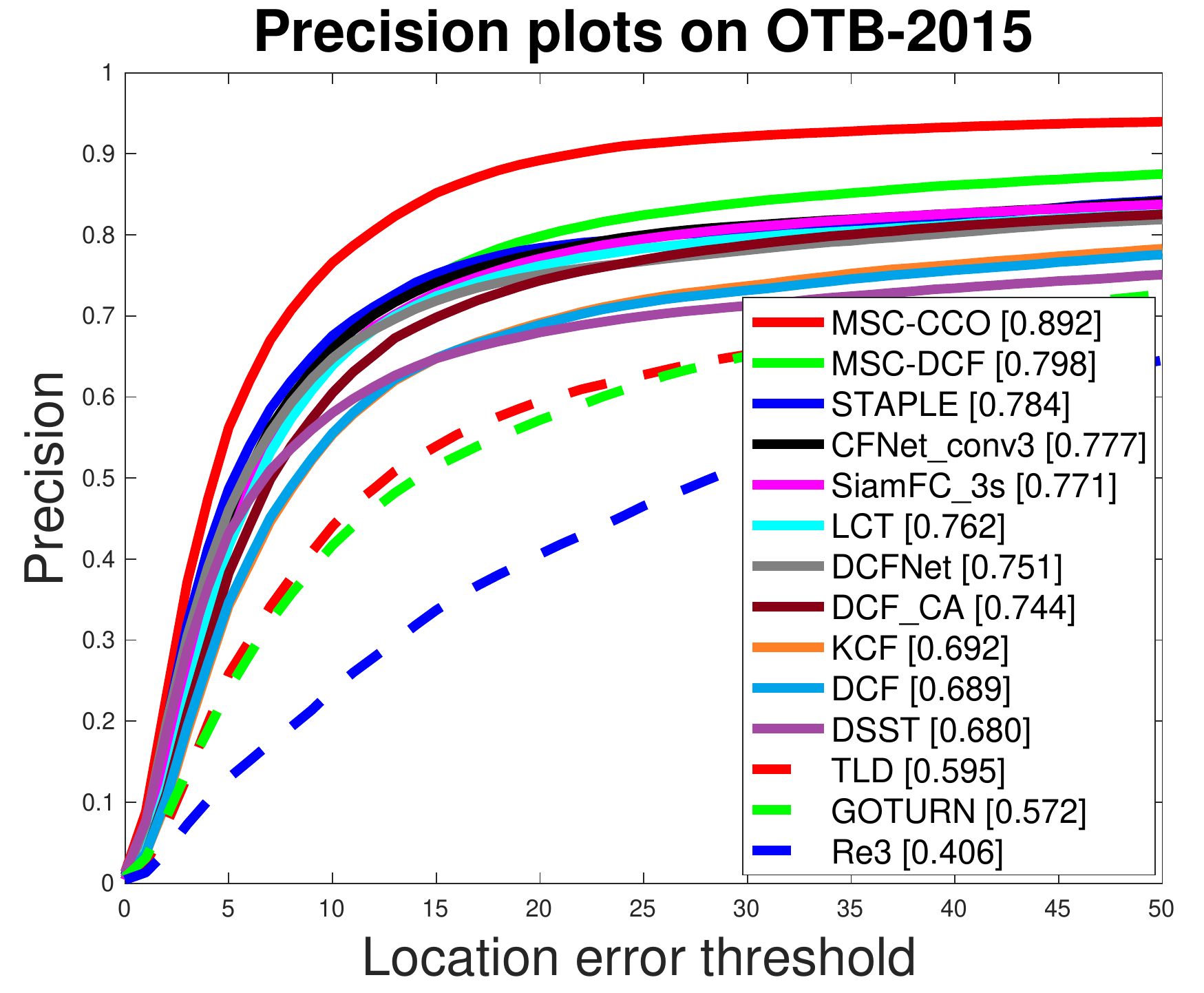}}
\end{minipage}}
%\vspace{-0.1cm}
 \subfigure{
\begin{minipage}[b]{.487\linewidth}
  \centerline{\includegraphics[width=5.98cm, height=4.5cm]{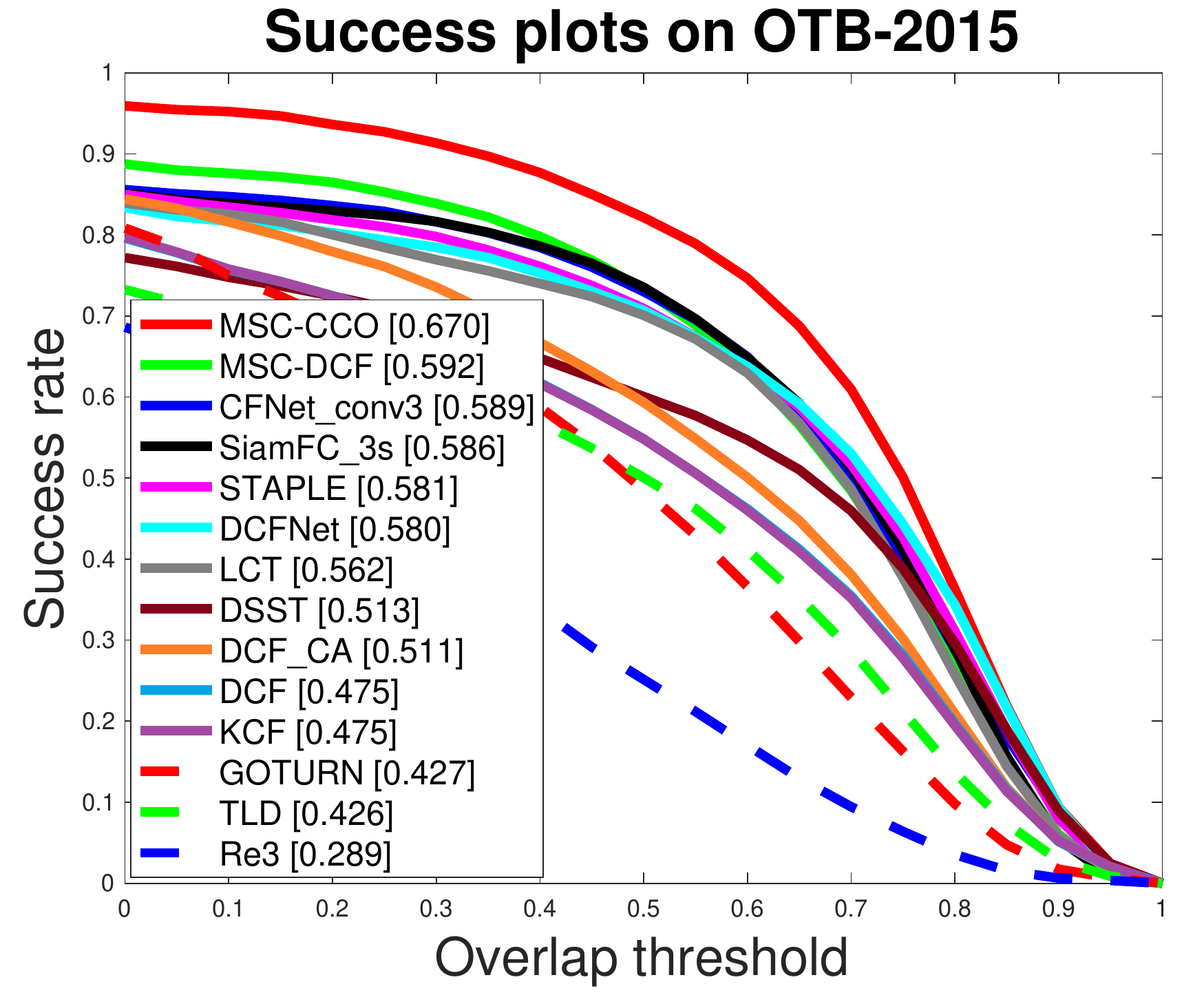}}
\end{minipage}}
\end{center}
%\vspace{-0.55cm}
\vfill
   \caption{\label{overall} Precision (left) and success (right) plots obtained by our MSC-DCF and MSC-CCO compared with the other 12 state-of-the-art real-time trackers on OTB-2015. DPRs and AUCs are reported in left and right brackets, respectively.}
\end{figure*}

%We compare the proposed tracker MSC-DCF with 11 state-of-the-art real-time trackers,
We compare the proposed highly efficient MSC-trackers (MSC-DCF and MSC-CCO) with 12 state-of-the-art trackers that can achieve real-time tracking speed (FPS$\textgreater$20) for fair comparison, including SiamFC \cite{SiamFC}, CFNet \cite{CFNet}, Staple \cite{Staple}, DCFNet \cite{DCFNet}, {DCF}$_{CA}$ \cite{CACF}, LCT \cite{LCT}, DSST \cite{DSST}, KCF \cite{KCF}, GOTURN \cite{GOTURN}, Re3 \cite{Re3}, DCF \cite{KCF}, and TLD \cite{TLD}.

%\textbf{Overall performance.} 
Fig. \ref{overall} compares the proposed MSC-trackers with the state-of-the-art real-time trackers, showing that our MSC-trackers achieve the best performance on OTB-2015. More specifically, MSC-CCO achieves the best DPR (89.2\%) followed by MSC-DCF (79.8\%) and CFNet (77.7\%). Note that CFNet is the winner of the VOT-17 real-time challenge. Similar to MSC-DCF, CFNet also employs the traditional DCF framework. However, different from CFNet,  our MSC-DCF learns complementary multi-layer features instead of single-layer features, thus achieving better performance than CFNet in terms of both DPR and AUC. 
%which demonstrates the effectiveness of the learned MSC features in accurate visual tracking. 

%Additionally, for the precision plot, the best DPR belongs to the propsoed MSC-DCF, followed by Staple (78.4), CFNet (77.7), and SiamFC (77.1). The Staple tracker incorporates both HOG and Color features in a DCF framework, achieving favorable tracking performance on the OTB dataset while maintaining high framerates. In comparison, our MSC-DCF outperforms it in both the accuracy and speed. 
\begin{table*}[!tp]
\newcommand{\tabincell}[2]{\begin{tabular}{@{}#1@{}}#2\end{tabular}}
\small
\begin{center}
%\vspace{-0.391cm}
\caption {\label{overallTable-MSCDCF}DPRs (\%) and speed ($^{*}$ indicates the GPU speed, otherwise the CPU speed) obtained by our MSC-DCF and MSC-CCO trackers as well as the state-of-the-art real-time trackers on OTB datasets. The results of top 8 performing trackers are given. The {\color{red} first},  {\color{green} second} and {\color{blue} third} best trackers are shown in color.}
%\label{table:headings}
\begin{tabular}{ccccccccc}
\hline\noalign{\smallskip}
&  \textbf{\tabincell{c}{MSC-CCO}}& \textbf{\tabincell{c}{MSC-DCF}} & \tabincell{c}{CFNet} &  \tabincell{c}{SiamFC} & \tabincell{c}{DCFNet}  & \tabincell{c}{Staple} & \tabincell{c}{DCF$_{CA}$} & \tabincell{c}{LCT}\\
\noalign{\smallskip}
\hline
\noalign{\smallskip}
\multirow{1}{*}{OTB-2013}
  & {\color{red}90.5}& {\color{blue}83.7} & 82.2 & 80.3 & 79.5 & 79.3 & 78.4 & {\color{green}84.8}   \\
%\hline
\multirow{1}{*}{OTB-2015}
 &{\color{red}89.2}& {\color{green}79.8} & 77.7 & 77.1 & 75.1 & {\color{blue}78.4} & 74.4 & 76.2 \\
 %\hline
 \multirow{1}{*}{OTB-50}
 &{\color{red}86.6}& {\color{green}75.2} & {\color{blue}72.3} & 69.4 & 68.3 & 68.1 & 71.2 & 69.1 \\
\hline
\hline
\multirow{1}{*}{Avg. DPR}
 &{\color{red}88.8}&{\color{green}79.6} & {\color{blue}77.4} & 75.6 & 74.3 & 75.3 & 74.7 & 76.7 \\
\multirow{1}{*}{Avg. FPS}
 &$20.6^*$&${66.8}^*$ & ${\color{blue}67.0}^*$ & ${\color{green}86.0}^*$ & $65.0^*$ & 48.3 & {\color{red}179.2} & 21.0 \\
\hline
%\multirow{1}{*}{Implemantation}
%& & M C & M C & M C & M C & M & M & M & M C & M C & M & M C & M C\\

\hline
\end{tabular}
\end{center}
\end{table*}

% reports the DPRs obtained by the competing trackers on OTB datasets, where t
As can be seen from Table \ref{overallTable-MSCDCF}, the proposed MSC-trackers achieve the best accuracy over all the three datasets. Specifically, MSC-CCO achieves the best accuracy (86.6\%) on OTB-50 followed by MSC-DCF (75.2\%). For the average DPR, the best two results belong to our MSC-trackers, followed by CFNet (77.4\%) and SiamFC (75.6\%). This comparison highlights the high accuracy achieved by our tracker among the state-of-the-art real-time trackers. The average tracking speed of different trackers are also reported in Table \ref{overallTable-MSCDCF}. Compared with other trackers, MSC-DCF achieves the fast tracking speed of 66.8 FPS while demonstrating its competitive tracking performance. In addition, higher accuracy of MSC-CCO is obtained at the cost of lower speed compared to MSC-DCF, but MSC-CCO still maintains quasi-real-time tracking speed of 20.6 FPS.

%Our second ranked MSC-DCF tracker significantly outperforms the third ranked CFNet tracker (72.3\%) with a large margin of 2.9\%.

%Higher accuracy of ECO, however, came at the cost of much lower speed compared to our MSC-CCO.
%accuracy performance. This empirically demonstrates the ability of our MSC features in efficient online object tracking.

%\vspace{0.95cm}

\subsection{Comparison with Deep Feature-based Trackers}
We compare the proposed MSC-trackers with 6 state-of-the-art deep feature-based CF trackers: CCOT \cite{CCOT}, MCPF \cite{MCPF}, CREST \cite{CREST}, DeepSRDCF \cite{DeepSRDCF}, CF2 \cite{HCF} and HDT \cite{HDT}.

%\textbf{Overall performance.} 
%Specifically, MCPF improves the searching strategy in CF2, utilizing many particle filters to search for the target. Despite its large searching region, MSC-CCO still achieves the better performance than MCPF, which demonstrates that the effectiveness of the learned MSC features.
%Additionally, the best AUCs on OTB-2013 and OTB-50 all belong to MSC-CCO, showing that our MSC-CCO can accurately track the target and estimate the target scale. 
\begin{figure*}[!tp]
    %\vspace{-0.1cm}
 \begin{center}
 \subfigure{
\begin{minipage}[b]{0.315\linewidth}
  \centerline{\includegraphics[width=4.15cm]{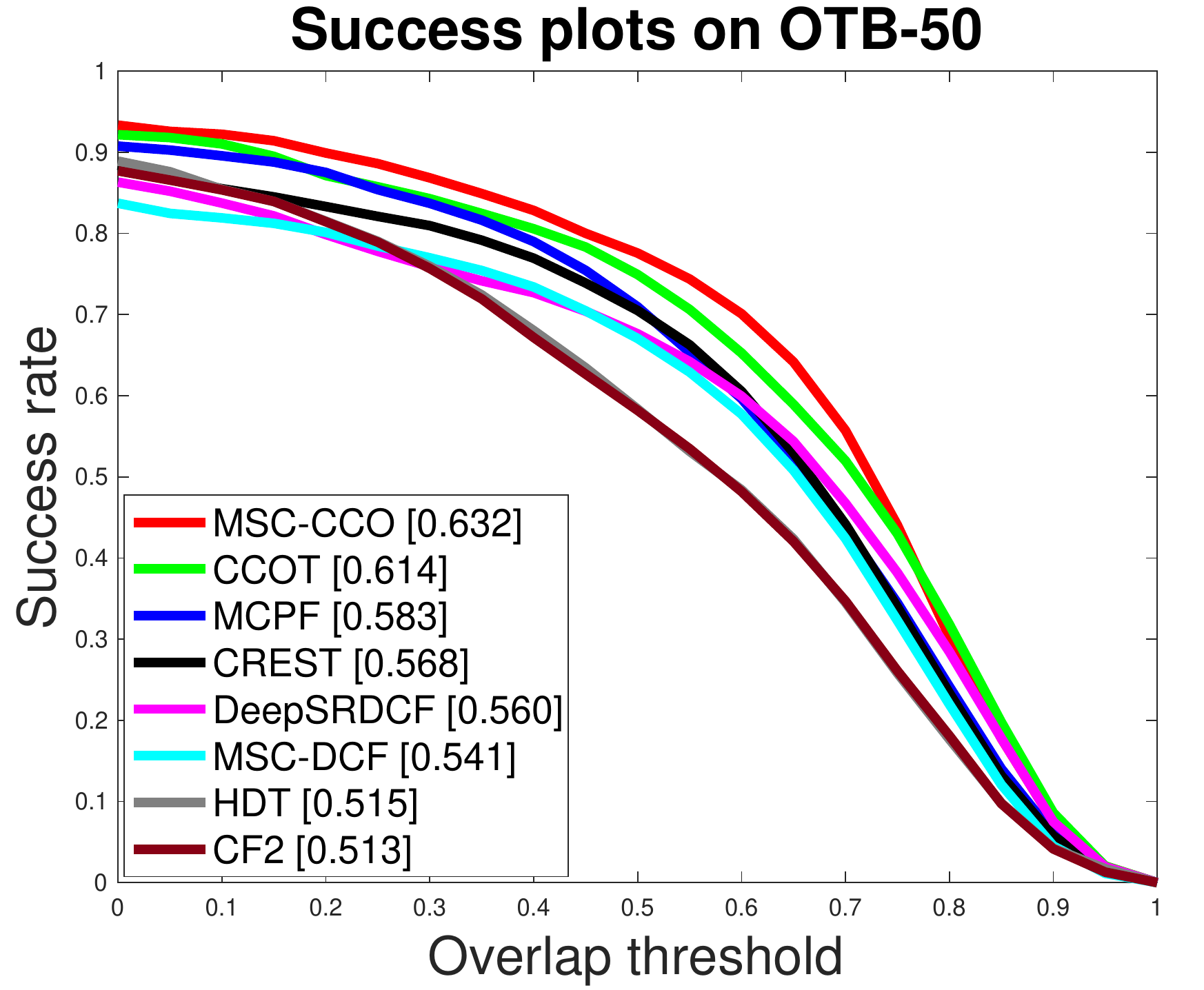}}
  \caption*{(a)}
\end{minipage}}
%\vspace{-0.1cm}
 \subfigure{
\begin{minipage}[b]{.315\linewidth}
  \centerline{\includegraphics[width=4.15cm]{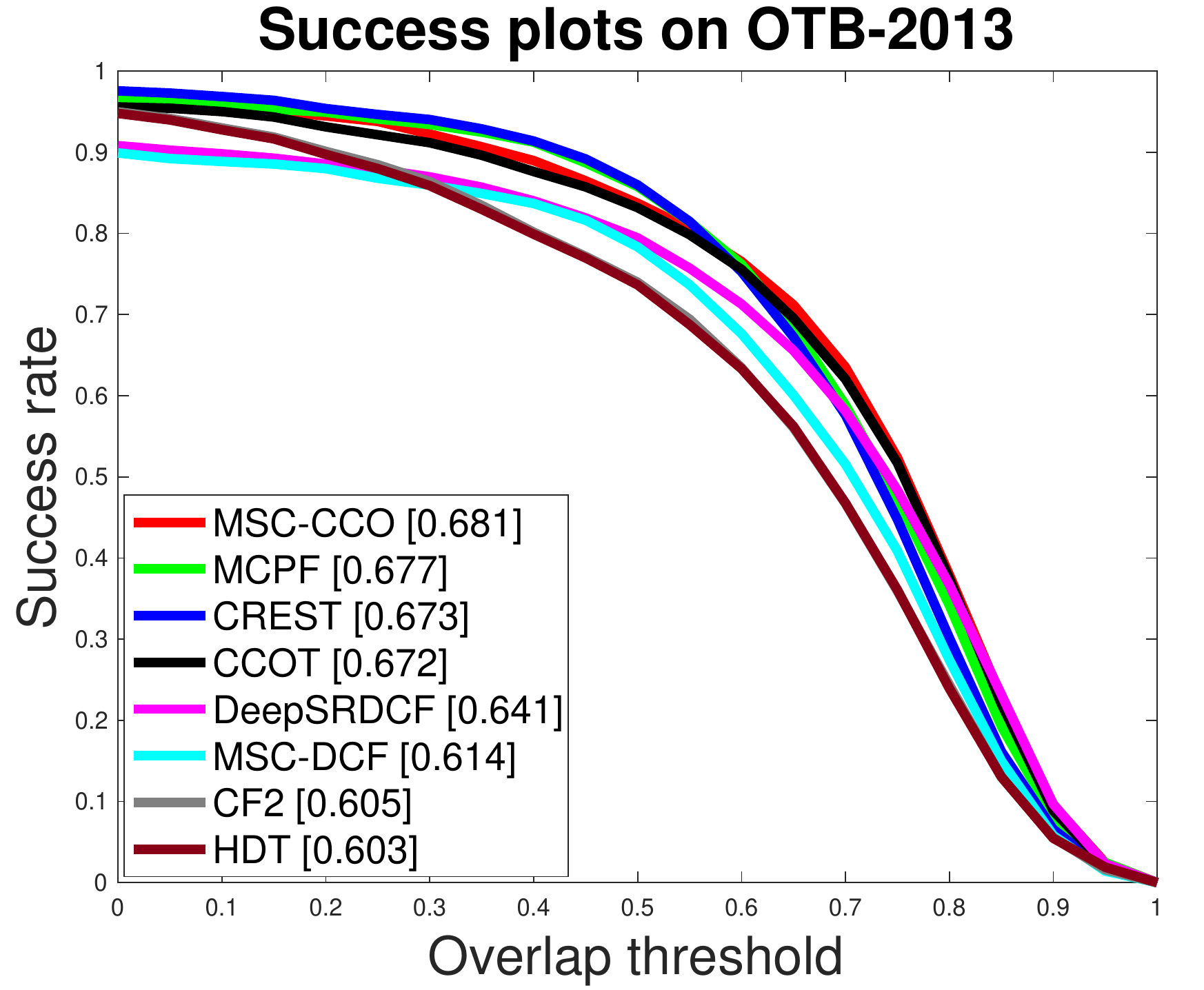}}
  \caption*{(b)}
\end{minipage}}
 \subfigure{
\begin{minipage}[b]{.315\linewidth}
  \centerline{\includegraphics[width=4.15cm]{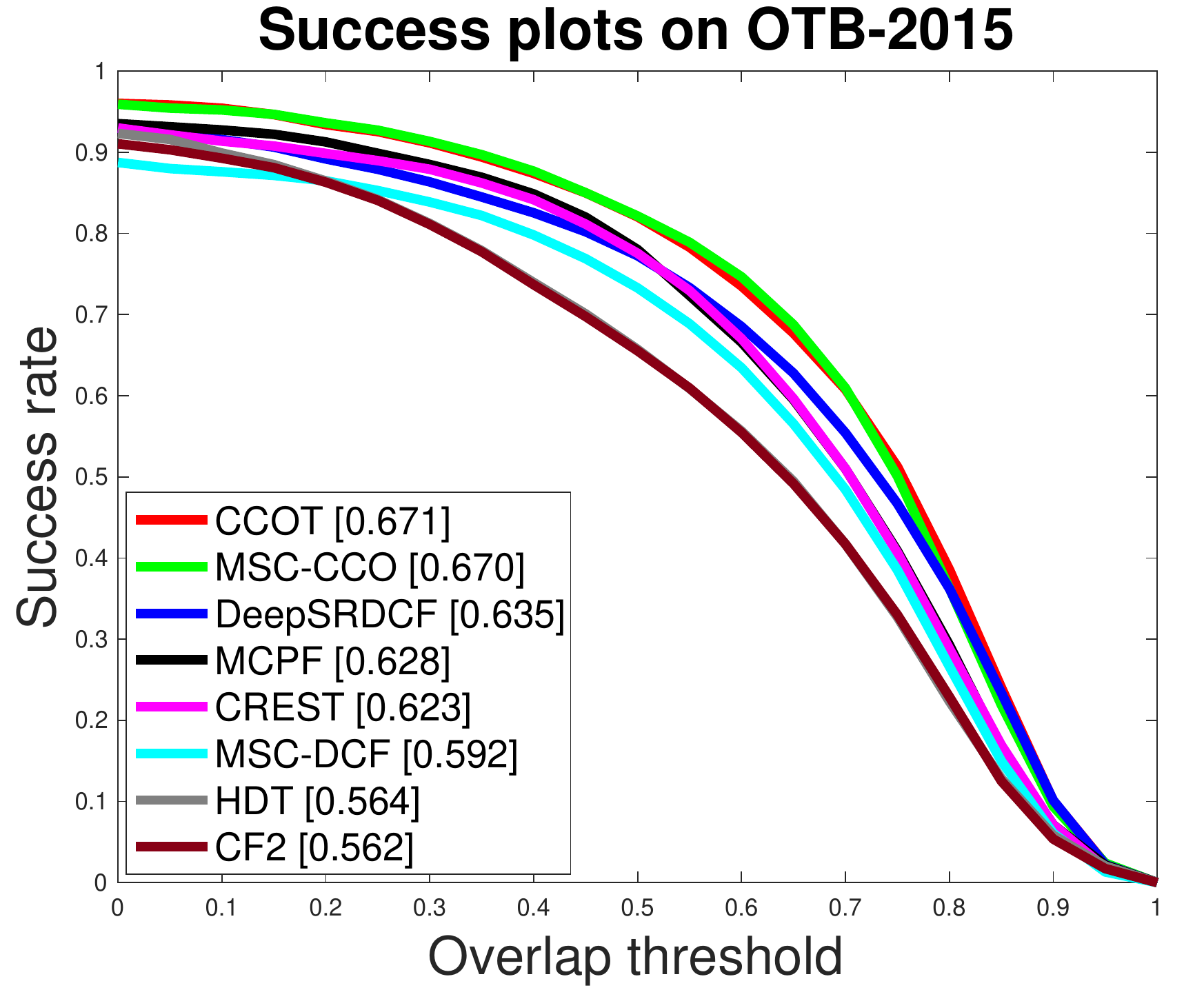}}
  \caption*{(c)}
\end{minipage}}
\end{center}
%\vspace{-0.55cm}
\vfill
   \caption{\label{MSC-CCO-overall} Success plots obtained by the proposed MSC-trackers (MSC-CCO and MSC-DCF) and the top-performing deep feature-based CF trackers on (a) OTB-50, (b) OTB-2013 and (c) OTB-2015.  AUCs are illustrated in brackets. } 
\end{figure*}

\begin{table*}[!tp]
\newcommand{\tabincell}[2]{\begin{tabular}{@{}#1@{}}#2\end{tabular}}
\small
\begin{center}
%\vspace{-0.391cm}
\caption {\label{overallTable-MSC-CCO}OSRs (\%) and speed ($^{*}$ indicates the GPU speed, otherwise the CPU speed) obtained by MSC-CCO and MSC-DCF as well as the state-of-the-art deep feature-based CF trackers on OTB-2013, OTB-2015 and OTB-50. The {\color{red} first},  {\color{green} second} and {\color{blue} third} best trackers are highlighted in color.}
%\label{table:headings}
\begin{tabular}{lcccccccc}
\hline\noalign{\smallskip}
&  \textbf{\tabincell{c}{MSC-CCO}} & \textbf{\tabincell{c}{MSC-DCF}} & \tabincell{c}{CCOT} &  \tabincell{c}{MCPF} & \tabincell{c}{DeepSRDCF}  & \tabincell{c}{HDT} & \tabincell{c}{CF2} & \tabincell{c}{CREST}\\
\noalign{\smallskip}
\hline
\noalign{\smallskip}
\multirow{1}{*}{OTB-2013}
   & {\color{blue}83.7} &78.3 &83.2 & {\color{green}85.8} & 79.4 & 73.7 & 74.0 & {\color{red}86.0}   \\
%\hline
\multirow{1}{*}{OTB-2015}
 & {\color{red}82.1} & 73.3&{\color{green}82.0} & {\color{blue}78.0} & 77.2 & 65.7 & 65.5 & 77.6 \\
 %\hline
 \multirow{1}{*}{OTB-50}
 & {\color{red}77.6} & 67.1&{\color{green}74.9} & {\color{blue}71.0} & 67.6 & 58.4 & 58.2 & 70.5 \\
\hline
\hline
\multirow{1}{*}{Avg. OSR}
 &{\color{red}81.1} & 72.9&{\color{green}80.0} & {\color{blue}78.3} & 74.7 & 65.9 & 65.9 & 78.0 \\
\multirow{1}{*}{Avg. FPS}
 &${\color{green}20.6}^*$ &${\color{red}66.8}^*$&$0.22^*$ & $0.56^*$ & \textless$1.0^*$ & ${\color{blue}11.1}^*$ & $10.5^*$ & $1.0^*$ \\
%\multirow{1}{*}{Implemantation}
%& & M C & M C & M C & M C & M & M & M & M C & M C & M & M C & M C\\
\hline
\end{tabular}
\end{center}
\end{table*}

Fig. \ref{MSC-CCO-overall} and Table \ref{overallTable-MSC-CCO} show the comparison of our MSC-trackers with several deep CF trackers with deep features on OTB datasets. More particularly, the AUC and OSR obtained by MSC-DCF are higher than those obtained by CF2 and HDT. MSC-CCO achieves the best OSR (77.6\%) on OTB-50, significantly outperforming the second best tracker CCOT with a large margin of 2.7\%. Furthermore, the average OSR obtained by MSC-CCO is 81.1\%, which is ranked at the first and followed by CCOT (80.0\%) and MCPF (78.3\%).  In terms of the tracking speed reported in Table \ref{MSC-CCO-overall}, compared with the other deep feature based trackers, MSC-DCF can run at 66.8 FPS, which is significantly faster than the compared trackers. In addition, MSC-CCO achieves the quasi-real-time tracking speed of 20.6 FPS, which is almost 94 times faster than CCOT and 37 times faster than MCPF. In the meanwhile, MSC-CCO achieves the best overall performance over all the three datasets. %This evaluation demonstrates that the learned MSC features are feasible for highly efficient visual tracking due to their compact and low-dimensional features.
%MSC-CCO not only achieves better performance, but also maintains the faster speed, 

% \begin{figure*}[!tp]
    %\vspace{-0.1cm}
% \begin{center}
 %\subfigure{
%\begin{minipage}[b]{0.48\linewidth}
 % \centerline{\includegraphics[width=6.1cm]{./figs/abla_1.pdf}}
%  \caption*{(a)}
%\end{minipage}}
%\vspace{-0.1cm}
 %\subfigure{
%\begin{minipage}[b]{.48\linewidth}
 % \centerline{\includegraphics[width=6.1cm]{./figs/abla_2.pdf}}
 % \caption*{(b)}
%\end{minipage}}
%\end{center}
%\vspace{-0.65cm}
%\vfill
 %  \caption{\label{ablation} DPRs obtained by MSC-DCF with respect to varying (a) the reserved shallow channel number K$_s$ and (b) the reserved deep channel number K$_d$ on OTB-2013. } 
%\end{figure*}

\begin{table*}[!tp]
\newcommand{\tabincell}[2]{\begin{tabular}{@{}#1@{}}#2\end{tabular}}
\small
\begin{center}
%\vspace{-0.391cm}
\caption {\label{ablation}DPRs (\%) obtained by MSC-DCF, MSC-CCO and their additional versions on OTB-2013 and OTB-2015.}

%MSC-DCF-CRM, MSC-CCO-CRM indicate MSC-DCF, MSC-CCO without using the CRM method, respectively.}

% w/o indicates without using CRM, otherwise indicates CRM is used. }
%\label{table:headings}
\begin{tabular}{c|c|c|c|c}
\hline\noalign{\smallskip}
&  {\tabincell{c}{MSC-CCO}} & {\tabincell{c}{MSC-CCO-w/o-CRM}}  &  {\tabincell{c}{MSC-DCF}}  & {\tabincell{c}{MSC-DCF-w/o-CRM}} \\
\noalign{\smallskip}
\hline
\noalign{\smallskip}
\multirow{1}{*}{OTB-2013}
   & 90.5 & 86.5 & 83.7 & 83.1    \\
\hline
\multirow{1}{*}{OTB-2015}
 & 89.2 &  87.1 & 79.8 & 79.2  \\
%\multirow{1}{*}{Implemantation}
%& & M C & M C & M C & M C & M & M & M & M C & M C & M & M C & M C\\
\hline
\end{tabular}
\end{center}
\end{table*}

\subsection{Ablation Study} %the reserved top ranked shallow channel numbers 
To demonstrate the effectiveness of the proposed CRM method, we evaluate MSC-DCF and MSC-CCO with additional versions, i.e., MSC-DCF and MSC-CCO without using the CRM method, namely MSC-DCF-w/o-CRM and MSC-CCO-w/o-CRM, respectively. As can be seen from Table \ref{ablation}, the CRM method is effective to improve tracking performance. Specifically, MSC-CCO achieves the accuracy (90.5\%) by improving 4\% of MSC-CCO-w/o-CRM on OTB-2013. For MSC-DCF, the performance is also improved by applying the CRM method. These results demonstrate the effectiveness of the CRM method, which is mainly due to the fact that the CRM method filters the feature channels that are more sensitive to the background regions while retaining the high-quality channels that are more beneficial for robust visual tracking. 

%In Fig. \ref{ablation}, we test different configurations of K$_{s}$ and K$_{d}$.  As can be seen, for the shallow channels, when K$_{s}$ decreases, the tracking performance will significantly decline. This is because the shallow features in the learned MSC features are compact, and they are important for the accurate localization of the target. Thus, for the shallow channels, we keep all the shallow channels produced by the proposed DSNet, i.e., K$_{s} = 32$. In addition, Fig. \ref{ablation}(b) shows that the optimal number of the reserved deep channels is 50. Further decreasing the reserved channel number K$_{d}$ will cause the performance degradation. In the meanwhile, reserving more deep channels (K$_{d}$ \textgreater 50) will introduce noisy information, which may affect the tracking performance. Thus, according to the proposed channel reliability measurement method, the top-50 ranked deep channels are selected for online tracking.

\section{Conclusions}
In this work, we propose a deep and shallow feature learning network (called as DSNet) to learn the multi-level same-resolution compressed (MSC) features, which effectively incorporate both deep and shallow features with the same spatial resolution for efficient online tracking. The proposed DSNet compresses multi-layer convolutional features and can be effectively trained in an end-to-end offline manner. The MSC features are generic and can be easily applied to any CF-based trackers. In addition, we propose an effective channel reliability measurement method to further refine the learned MSC features. To demonstrate the effectiveness of our MSC features, two MSC features based trackers are presented, namely MSC-DCF and MSC-CCO, respectively. Experiments on several large scale benchmarks show that the proposed methods perform favorably against state-of-the-art tracking methods.
\\

\noindent \textbf{Acknowledgments.} This work is supported by the National Natural Science of China (Grant No. U1605252, 61872307, 61472334 and 61571379) and the National Key Research and Development Program of China under Grant No. 2017YFB1302400.

% MSC-DCF outperforms several state-of-the-art real-time trackers while running at 66.8 FPS, and our MSC-CCO achieves superior performance in comparison to several top-performing deep feature-based trackers in terms of both tracking accuracy and speed.

%This work was supported by the National Natural Science Foundation of China under Grants U1605252, 61472334 and 61571379.

%\subsection{ACCV Paper Submission Notes}
%Papers accepted for the ACCV conference will be allocated 16 pages (maximum 14 pages for content plus maximum two pages for references) in the proceedings. 

\bibliographystyle{splncs04}
\bibliography{egbib}
%Author, F.: Article title. Journal \textbf{2}(5), 99--110 (2016)

%Author, F., Author, S.: Title of a proceedings paper. In: Editor,
%F., Editor, S. (eds.) CONFERENCE 2016, LNCS, vol. 9999, pp. 1--13.
%Springer, Heidelberg (2016). \doi{10.10007/1234567890}

%Author, F., Author, S., Author, T.: Book title. 2nd edn. Publisher,

%Author, A.-B.: Contribution title. In: 9th International Proceedings
%on Proceedings, pp. 1--2. Publisher, Location (2010)

%LNCS Homepage, \url{http://www.springer.com/lncs}. Last accessed 4
%\end{thebibliography}
\end{document}